\documentclass[final,sglblindworkshop]{article} 
\usepackage{neurips_2025}

\usepackage[utf8]{inputenc} 
\usepackage[T1]{fontenc}    
\usepackage{amsmath} 
\usepackage{hyperref}       
\usepackage{url}            
\usepackage{booktabs}       
\usepackage{amsfonts}       
\usepackage{nicefrac}       
\usepackage{microtype}      
\usepackage{xcolor}         
\usepackage[pdftex]{graphicx}
\graphicspath{{./images/}}

\newcommand{\R}{\mathbb{R}}

\newcommand{\N}{\mathbb{N}}

\newcommand{\E}{\mathbb{E}}

\definecolor{biascolor}{rgb}{0.85, 0.95, 1.0}  

\newcommand{\biascorr}[1]{%
  \mathord{\colorbox{biascolor}{\hbox{$\displaystyle #1$}}}%
}
\newcommand{\biastext}[1]{%
  \colorbox{biascolor}{#1}}%

\workshoptitle{EurIPS 2025 Workshop on Principles of Generative Modeling (PriGM).}
\title{Simplifying Adam: Bias Correction Debunked}

\author{
  Sam Laing \\
ELLIS Institute T\"ubingen, MPI-IS\\
  T\"ubingen AI Center, Germany\\ 
  \texttt{slaing155@gmail.com}
  \And
  Antonio Orvieto \\
    ELLIS Institute T\"ubingen, MPI-IS\\
  T\"ubingen AI Center, Germany\\\\
}

\begin{document}
\maketitle

\begin{abstract}
    The Adam optimizer is a cornerstone of modern deep learning, yet the empirical necessity of each of its individual components is often taken for granted. This paper presents a focused investigation into the role of bias-correction, a feature whose contribution remains poorly understood. Through a series of systematic ablations on vision and language modelling tasks, we demonstrate that the conventional wisdom surrounding bias correction is misleading. In particular, we demonstrate that in the optimal hyper-parameter configuration, the inclusion of bias correction leads to no improvement in final test performance. Moreover, unless appropriate learning rate scheduling is implemented, the inclusion of bias correction can sometimes be detrimental to performance.  We further reinterpret bias correction as a form of implicit learning rate scheduling whose behaviour is strongly dependent on the choice of smoothing hyperparameters $\beta_1, \beta_2 \in [0,1)$. Our findings challenge the universal inclusion of this component. 
\end{abstract}
\section{Introduction}
The Adam optimizer \cite{kingma2017adammethodstochasticoptimization} (with decoupled weight decay \cite{loshchilov2019decoupledweightdecayregularization}) has established itself as the de facto standard in deep learning. Due to its robust empirical performance and ease of implementation, it is commonly adopted as the default choice when training neural networks both large and small and across a wide range of tasks. Practitioners often use Adam without extensive hyper-parameter tuning or consideration of each of the individual components of the optimizer. One such component is \emph{bias correction}. \\
A single step of the Adam optimizer, with \biastext{bias correction}, is given by 
\begin{equation}
\label{eq: full adam}
\begin{aligned}
    m_{t} &= \beta_1 m_{t-1} + (1-\beta_1)g_{t} \\
    v_{t} &= \beta_2 v_{t-1} + (1-\beta_2) g_{t}^2\\
    \hat{m}_{t} &= \displaystyle \biascorr{\frac{1}{1-\beta_1^t}} m_t \ , \ 
    \hat{v}_{t} = \displaystyle \biascorr{\frac{1}{1-\beta_2^t}} v_t  \\
    \theta_t &= \theta_{t-1} - \eta_t \frac{\hat{m}_t}{\sqrt{\hat{v}_t} + \varepsilon}
\end{aligned}
\end{equation}
where $g_t$ is the stochastic gradient, $m_t, v_t$ are the respective first and second moments (which are typically initialized at $0$),  $(\beta_1, \beta_2) \in [0, 1)^2 $ are the exponential decay rates, $\lambda \in \R_{\ge 0}$ weight decay and $(\eta_j )_{j\ge 1} \subset \R_{\ge 0}$ is the learning rate schedule. 
\paragraph{Background and Related Work}
While bias correction is universally included in practical implementations of Adam(\cite{jax2018github}, \cite{paszke2019pytorchimperativestylehighperformance}, \cite{tensorflow2015-whitepaper}), it is inconsistently treated in the theoretical literature. In many analyses, the term is explicitly incorporated (\cite{balles2020dissectingadamsignmagnitude}, \cite{li2023convergenceadamrelaxedassumptions}), but is sometimes ignored for simplification (\cite{bernstein2024oldoptimizernewnorm}) . To the best of our knowledge, it has not been carefully ablated in empirical studies across a range of hyperparameter configurations and tasks.  Nevertheless, it plays a non-trivial role in shaping the optimizer's behaviour and, when included, tends to complicate convergence analyses and add interpretive nuance. For example, \cite{defossez2022simpleconvergenceproofadam} discuss its influence in the context of second-moment estimation. Explicitly, the authors omit the correction term for the first moment $m_t$, but not $v_t$ arguing that this "simplifies the analyses". Moreover, nearly all existing analyses assume near default settings of $\beta_1 = 0.9, \beta_2 = 0.999$ or $0.99$, leaving unexplored how the effects of bias correction might change across the $(\beta_1, \beta_2)$-landscape. This gap in the literature is particularly relevant in light of recent findings, such as \cite{orvieto2025searchadamssecretsauce}, which demonstrate that the setting $\beta_1 = \beta_2$ generally achieves optimal performance when pretraining large language models. 
\paragraph{A Discussion of the ``Proof'' for Bias Correction} 
One can express the moments $m_t, v_t$ at time step $t\in \N$ in closed-form as follows:
\begin{equation}
    m_t = (1-\beta_1)\sum_{j=1}^{t} {\beta_1^{t-j}g_j} \quad, \quad v_t = (1-\beta_2)\sum_{j=1}^{t} {\beta_2^{t-j}g_j^2}
\end{equation}
The following argument is used by the original authors to justify the inclusion of the bias correction step
\begin{align*}
	\E[m_t] &= \E\left[ (1-\beta_1)\sum_{i=1}^{t} {\beta_1^{t-i} g_i}  \right]  \\
		&= \E\left[(1-\beta_1)g_t \sum_{i=1}^{t} {\beta_1^{t-i}}\right] \quad (\text{assuming $\E[g_t] \approx E[g_i]$ for $i<t$)}\\
		&= (1-\beta_1^{t})\E[g_t]
\end{align*}
It is therefore argued that dividing by  $1-\beta_1^{t}$ removes the expected "bias" in the exponential moving average. An analogous argument is used to justify the bias correction factor for $v_t$. \\
While the assumption that $\E[g_t] \approx \E[g_i]$ for $i<t$ at early training steps simplifies the analysis, it usually does not hold in practice. In particular, unless a very gradual warm-up of learning rate is applied, it is unlikely that stepping somewhere in a complex loss landscape would not cause a significant change of the expected value of the gradient. 
\paragraph{Contributions} In this work, we challenge the conventional understanding of bias correction. Through controlled ablations in the language and vision settings, we show that:
\begin{itemize}
    \item  The inclusion of bias correction induces an \emph{implicit learning rate schedule} by altering the effective learning rate.
    \item For the $\beta_1=\beta_2$ setting (LLM-optimal), bias correction provides no benefit and can even degrade performance unless appropriate learning rate scheduling is implemented.
    \item For default parameters where performance is suboptimal, its removal is detrimental, explaining the source of conventional wisdom. 
\end{itemize}

\section{Experiments}
\begin{figure}[t]
    \centering
\includegraphics[width=0.9\linewidth]{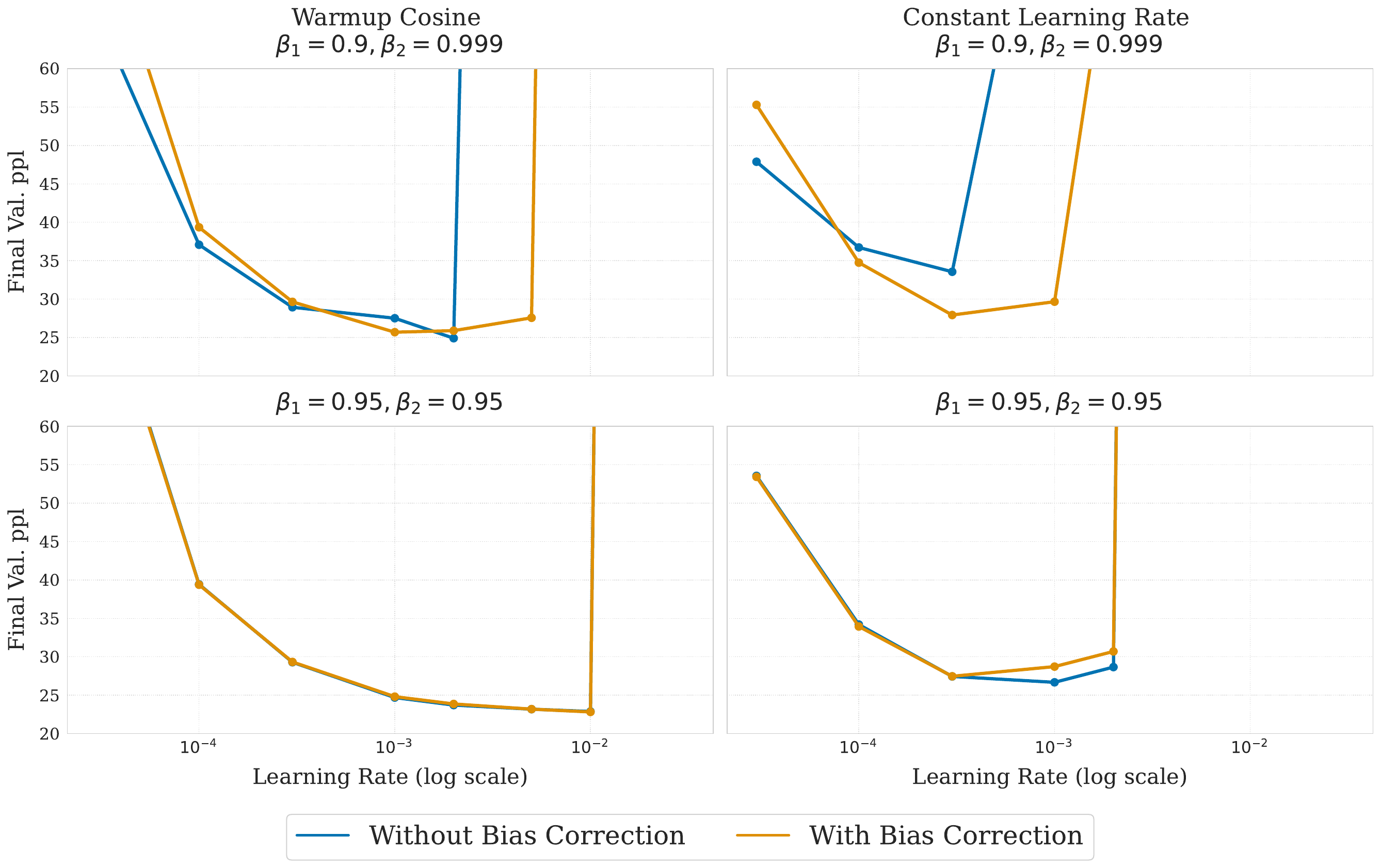}  
    \caption{
    Sensitivity to learning rate for AdamW with and without bias correction (orange and blue respectively). The plots show final validation perplexity (y-axis) across a range of learning rates (x-axis, log-scale). Results are averaged over $3$ random seeds.\\
    With warm-up cosine scheduling, removing bias correction increases sensitivity for default hyperparameters $(\beta_1,\beta_2)=(0.9,0.999)$ but with identical optimal performance. For the LM-optimal setting $(\beta_1, \beta_2) = (0.95, 0.95)$, performance is identical.
    \\
    With a fixed learning rate, the inclusion of bias correction has a more pronounced effect. In the default torch setting $(0.9, 0.999)$, excluding bias correction has a detrimental effect whereas for the LM-optimal setting $(0.95, 0.95)$, bias correction slightly degrades optimal performance. 
    }
    \label{fig: ppl_panel999}
\end{figure}
\begin{figure}[t]
    \centering
\includegraphics[width=\linewidth]{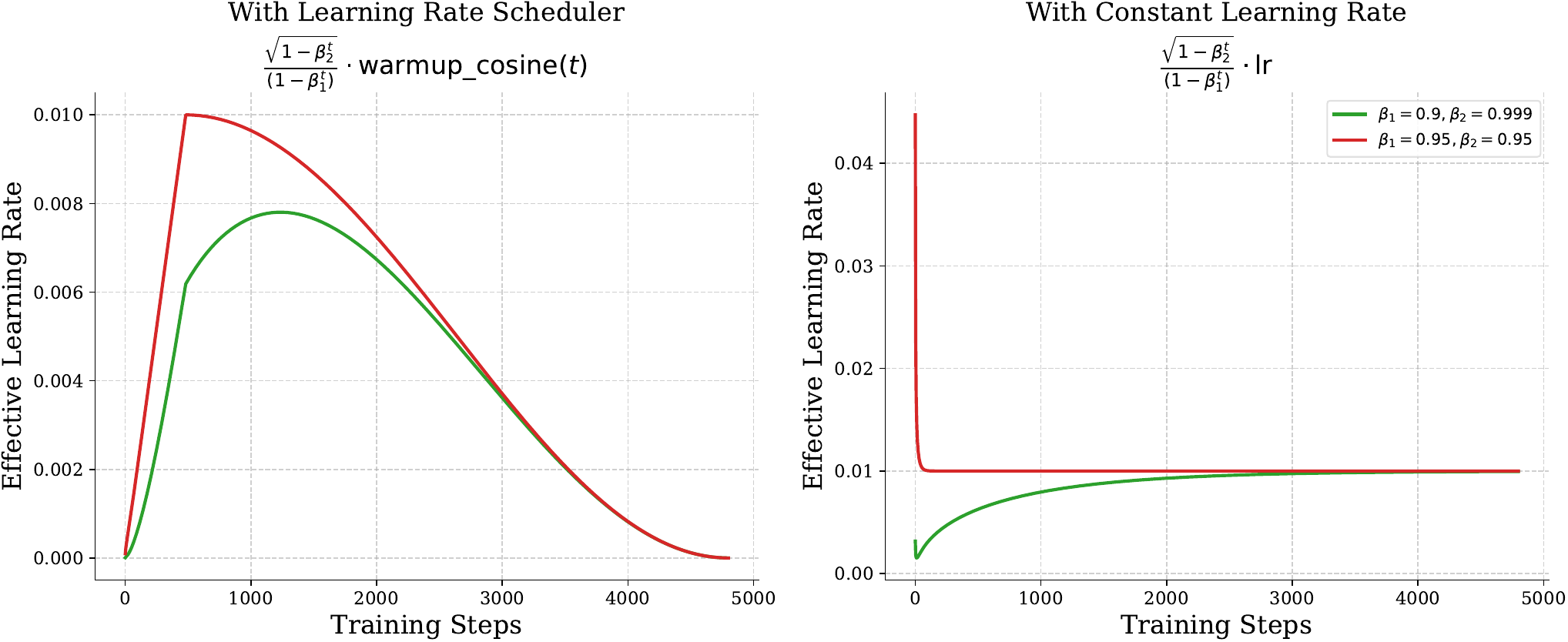}  
    \caption{
        Comparison of the effective learning rate when bias correction is applied for $(\beta_1, \beta_2) = (0.9, 0.999)$ (green) and $(\beta_1, \beta_2) = (0.95, 0.95)$ (red) under both warm-up cosine scheduling (left) and a constant learning rate (right).\\
        With warm-up cosine scheduling, the bias correction factor is effectively absorbed for the LM-optimal setting $(0.95, 0.95)$ (the true warmup cosine schedule is indistinguishable from the red curve), whereas for the default setting $(0.9, 0.999)$ it substantially modifies the effective learning rate, lowering the peak value.  
        Without scheduling, the torch default configuration exhibits a very gradual warm-up on effect on the effective learning rate, while the LM-optimal setting produces an initial spike that quickly decays to the nominal learning rate. 
     }
    \label{fig: effective_lr4800}
\end{figure} 
\paragraph{Language Model Training Details}
In the case of pretraining language models, we focus exclusively on transformer-based models. Specifically, we make use of an enhancement (\cite{ajroldi2024plainlm}) of nanoGPT (\cite{Karpathy2022}), which incorporates improvements such as RMSNorm  \cite{zhang2019rootmeansquarelayer} in place of batch/layer normalization, SwiGLU \cite{SwiGLU}, a FlashAttention \cite{dao2022flashattentionfastmemoryefficientexact} mechanism and Rotary Positional Embeddings (\cite{su2023roformerenhancedtransformerrotary}).
\\
All language models were trained on the SlimPajama \cite{shen2024slimpajamadcunderstandingdatacombinations} dataset. We pretrain a $160$M parameter, $12$ layer model with hidden size $768$ on $2.5$B tokens, varying the learning rate and investigating a number of different $(\beta_1, \beta_2)$ settings. In all runs, we apply decoupled weight decay\cite{loshchilov2017sgdrstochasticgradientdescent} and global gradient clipping \cite{zhang2020adaptivemethodsgoodattention} for optimal performance\footnotemark{}\footnotetext{The dynamics without decoupled weight decay/gradient clipping are nearly identical but with generally worse performance throughout runs. } . The batch size is held fixed at $256$.
\\
We further compare the performance when a warm-up cosine schedule (linear warm-up for the first $10$\% of steps followed by cosine decay to zero) is applied vs a fixed learning rate throughout training. 
\\
We compare the performance when training in the torch default setting $(\beta_1, \beta_2) = (0.9, 0.999)$ and the more language model pretraining-optimal setting $\beta_1 = \beta_2 = 0.95$ which was shown in \cite{orvieto2025searchadamssecretsauce} to yield near best performance across a large sweep. 
Figure \ref{fig: ppl_panel999} displays the final validation perplexity across a sweep of learning rates for the two settings. 
In Appendix \ref{sec: b1isb2}, we extend our investigation of the $\beta_1 = \beta_2$ case, systematically evaluating different values of this shared parameter (see Figure \ref{fig: ppl_panel} and an explanation in Figure \ref{fig: b1_is_b2}). 
\paragraph{Vision Models}
To support our intuition, we perform further experiments in the vision setting. Details, experimental results and discussion are all provided in the appendix \ref{sec: vision}. In particular, we display our empirical findings in Figures \ref{fig:resnet9_cifar10}, \ref{fig:vit_tinyimgnet}, \ref{fig:resnet50_tinyimgnet}.
\section{Bias Correction as Implicit Learning Rate Scheduling}
The bias-corrected Adam step direction (ignoring $\varepsilon$) factorizes as:
\begin{equation}
	\frac{\hat{m}_t}{\sqrt{ \ \hat{v}_t}} = \cfrac{\frac{1}{1-\beta_1^{t}} m_t}{\sqrt{ \frac{1}{1-\beta_2^{t}}v_t}} = \cfrac{\sqrt{1-\beta_2^{t}} }{1-\beta_1^{t}} \ \cfrac{m_t}{\sqrt{v_t} } = \rho(t; \beta_1, \beta_2) \frac{m_t}{\sqrt{v_t} }
	\label{eq: bias factor}
\end{equation}
Where we define the \textit{bias-correction factor} $\rho(t;\beta_1,\beta_2) := \frac{\sqrt{1-\beta_2^{t}} }{1-\beta_1^{t}}$. This term modulates the \textit{effective learning rate} $\rho(t; \beta_1, \beta_2) \cdot \eta_t$ over time in a manner which depends heavily on the values of $\beta_1, \beta_2$. \\
While prior work \cite{john2021adamdimprovedbiascorrectionadam}  \cite{defossez2022simpleconvergenceproofadam} has recognized this interpretation of the bias-correction factor, its empirical behaviour has only been considered for a limited subset of $(\beta_1, \beta_2)$ \cite{kingma2017adammethodstochasticoptimization} -- typically those near the default settings ($\beta_1 = 0.9, \beta_2 = 0.999$). 
\\
Figure \ref{fig: effective_lr4800} reveals how $\rho(t; \beta_1, \beta_2)$ behaves differently across configurations:
\begin{itemize}
    \item \textbf{Default setting} $(0.9, 0.999)$: Creates a very gradual warm-up effect, slowly increasing the effective learning rate across thousands of iterations. 
    \item \textbf{LM-optimal setting} $(0.95, 0.95)$: Produces a large initial spike that quickly decays to baseline
\end{itemize}

The interaction with explicit scheduling is crucial. Warm-up cosine scheduling completely absorbs the $\rho(t)$ spike for $(0.95, 0.95)$, but the bias correction factor non-negligibly alters the effective learning rate for $(0.9, 0.999)$. This results in a dampened peak learning rate. 
\\
This explains our empirical results in Figure \ref{fig: ppl_panel999}:
\begin{itemize}
    \item With scheduling, performance differences vanish for $\beta_1 = \beta_2$ as the spike is absorbed
    \item Without scheduling, the $(0.95, 0.95)$ spike causes instability in the early training steps, degrading performance
    \item Default parameters benefit from bias correction's implicit warm-up when no scheduling is used. 
\end{itemize}

\section{Conclusion}
We have demonstrated a clear and actionable finding: when pretraining language models with a proper learning rate schedule and optimal hyperparameters, Adam achieves the same validation performance with or without the inclusion of bias correction. Bias correction is not a true performance enhancer; but merely an implicit, and often clumsy, learning rate warm-up. Since explicit learning rate scheduling is always required to achieve optimal results, we therefore advocate for its removal from both implementation and theoretical analysis. This ultimately yields a simpler, more interpretable optimizer without sacrificing performance. It also removes a potential confounder in convergence studies and large scale training. 

\section*{Acknowledgements}

 The authors acknowledge the financial support of the Hector Foundation, and are thankful for the compute resources made available by MPI-IS and the Tübingen AI ecosystem. We additionally thank Philipp Hennig for the comments and remarks on our evaluations
\bibliographystyle{plainnat}

\bibliography{bibliography}
\newpage
\appendix
\section{Additional Experiments in the Vision Setting}
\label{sec: vision}
\begin{figure}[ht]
    \centering
    \includegraphics[width=\linewidth]{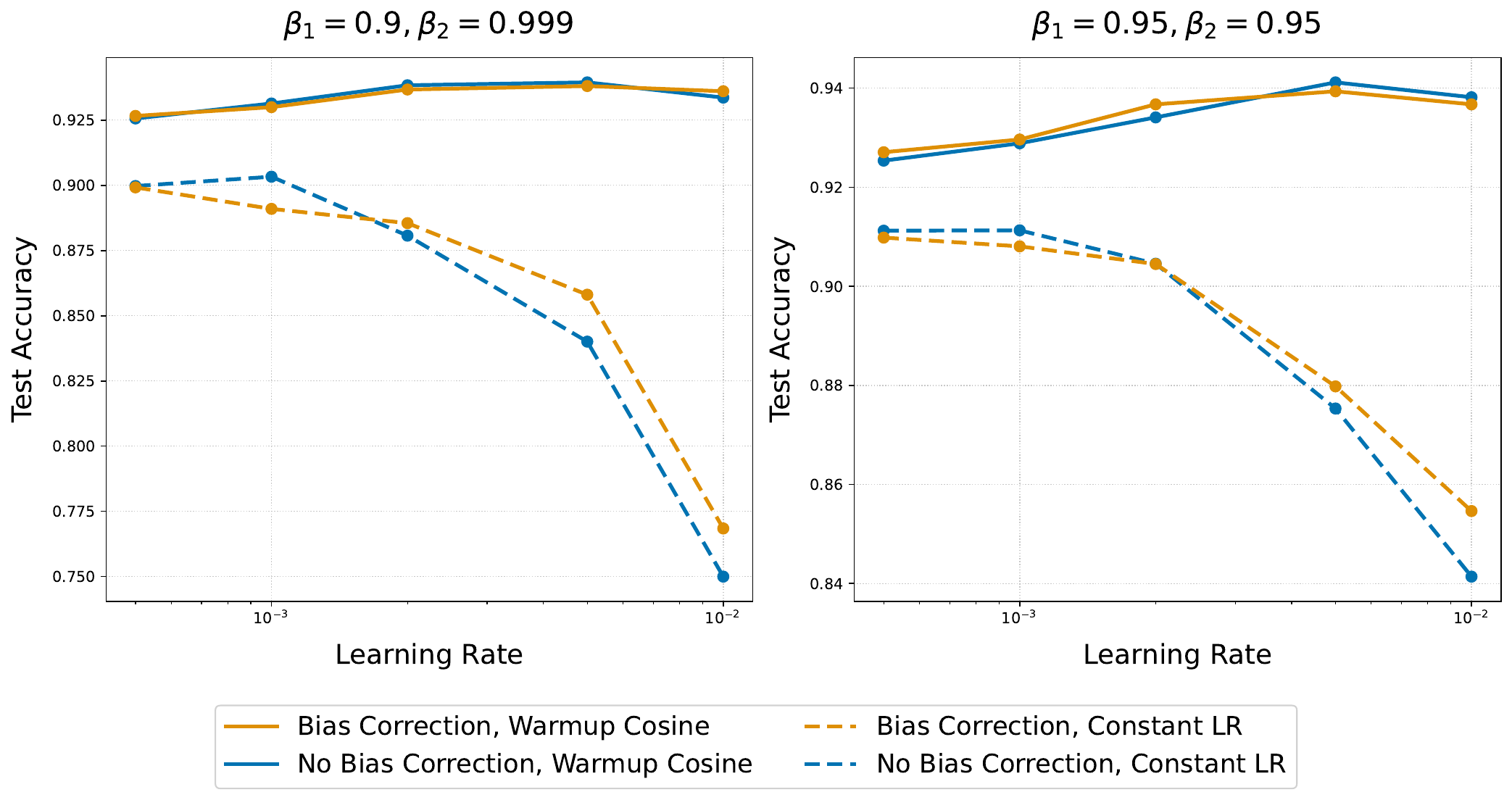}
    \caption{ResNet9 on CIFAR-10.}
    \label{fig:resnet9_cifar10}
\end{figure}

\begin{figure}
\centering
    \includegraphics[width=\linewidth]{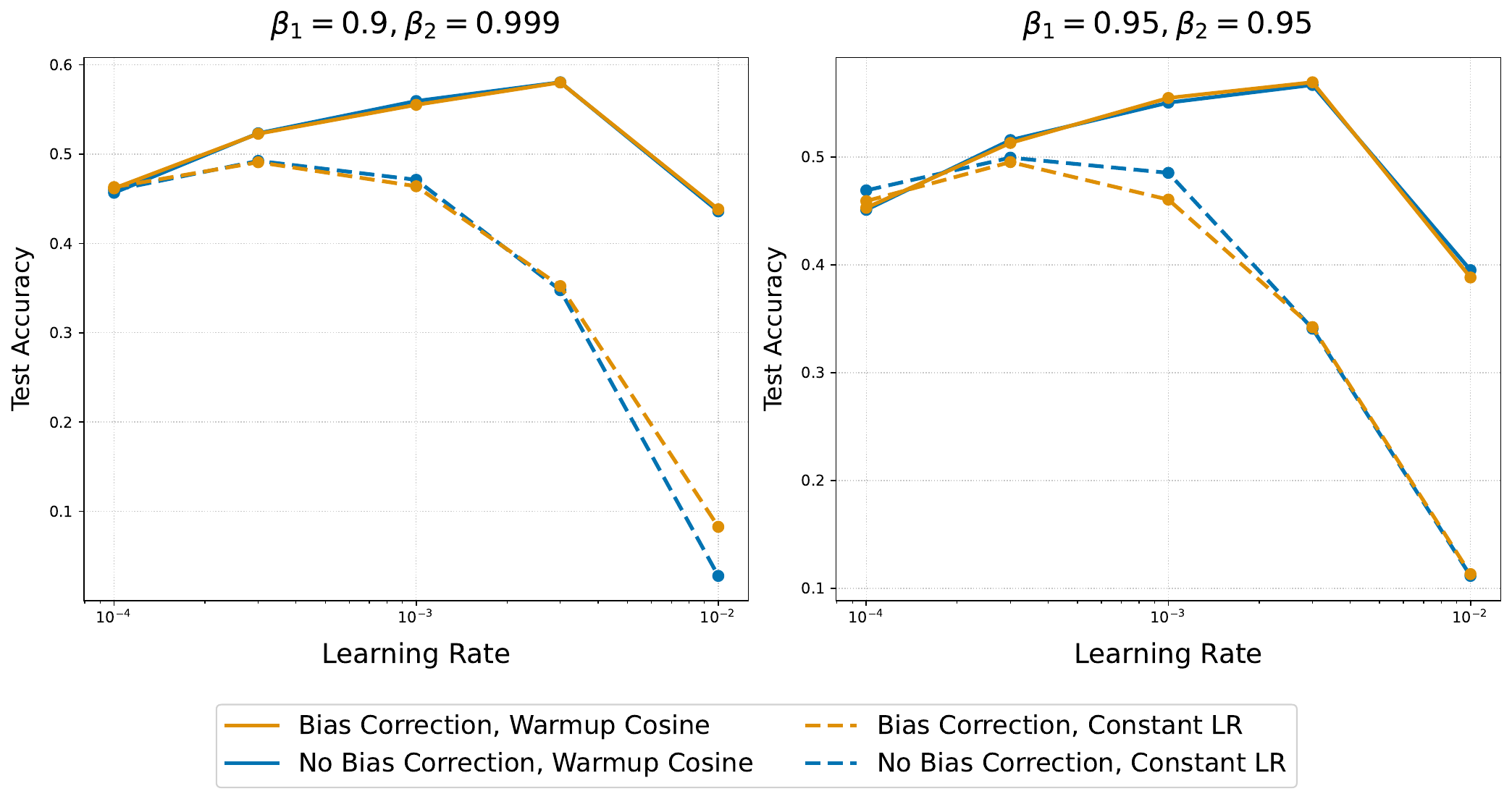}
    \caption{ResNet50 on Tiny ImageNet.}
    \label{fig:resnet50_tinyimgnet}
\end{figure}

\begin{figure}
\centering
    \includegraphics[width=\linewidth]{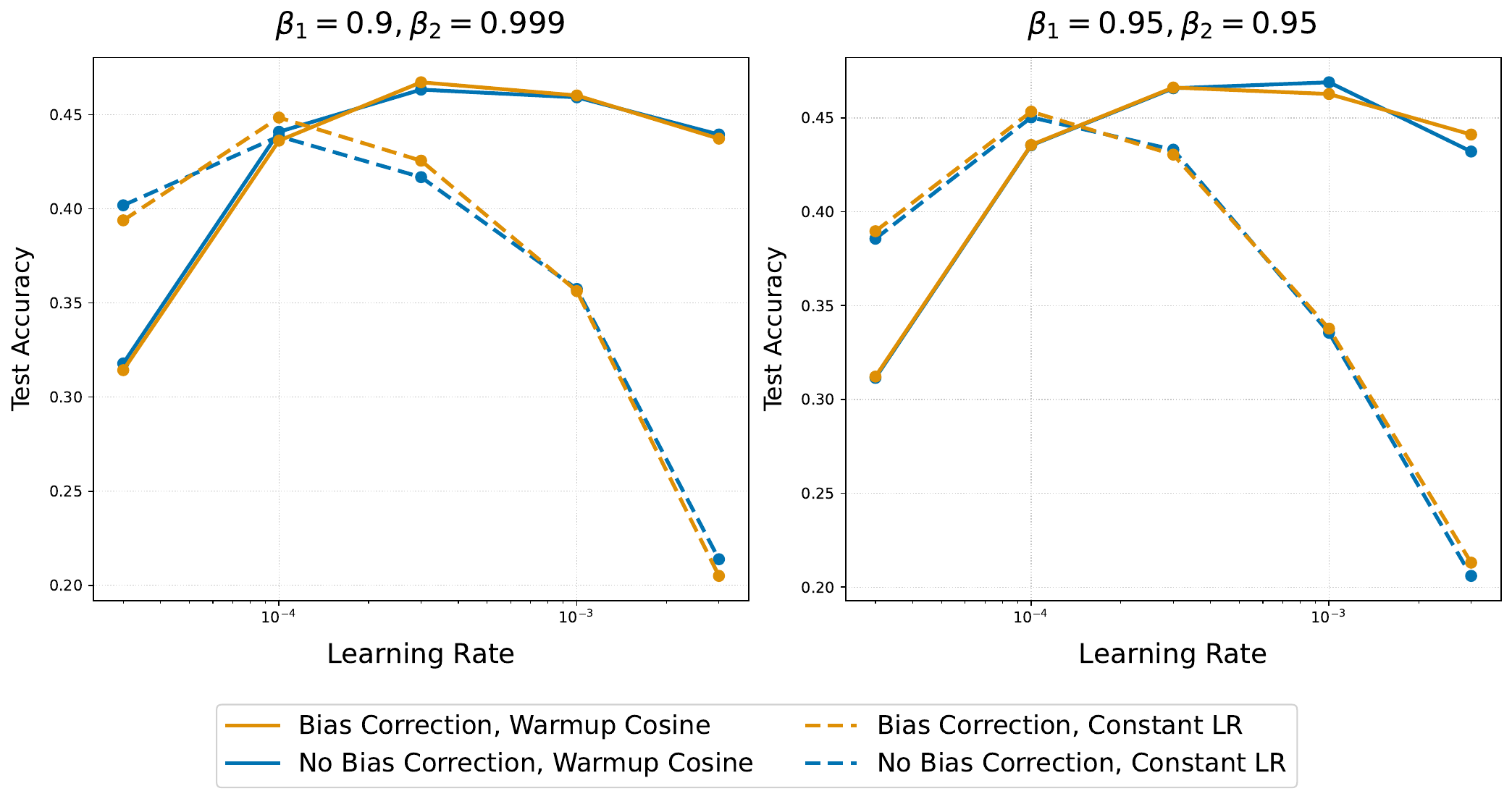}
    \caption{ViT on Tiny ImageNet}
    \label{fig:vit_tinyimgnet}
\end{figure}

\paragraph{Vision Training Setup}
In order to support our claims, we investigate the effect of removing bias correction in the vision setting. We train a ResNet9\footnotemark{} model on CIFAR-10 \cite{krizhevsky2009learning} and both a vision transformer (ViT) \cite{dosovitskiy2021imageworth16x16words} and ResNet \cite{he2015deepresiduallearningimage} model on TinyImagenet \cite{hendrycks2019benchmarking}. Standard data augmentation was applied in all cases along with decoupled weight decay \cite{loshchilov2017sgdrstochasticgradientdescent}. We compare the effect in the torch default setting $(\beta_1, \beta_2) = (0.9, 0.999)$ and the $\beta_1=\beta_2 = 0.95$ setting. Again we consider both the case of warm-up cosine scheduling and constant learning rate. For each learning rate, we compute the average test accuracy from 3 random seeds.  \\
\footnotetext{A lightweight ResNet implementation \cite{he2015deepresiduallearningimage} with $\sim 1$M parameters.  }
\paragraph{Experimental Results} Across the three model-dataset settings (Figure \ref{fig:resnet9_cifar10}, Figure \ref{fig:vit_tinyimgnet}, \ref{fig:resnet50_tinyimgnet}, we observe no qualitative difference in test performance when removing bias correction. This is true for both the warm-up cosine and constant learning rate cases and for both $(\beta_1, \beta_2)$ settings. 
\\
\paragraph{Takeaway} These results further support our claims that bias correction can be safely removed from Adam without decreased performance, provided a solid training pipeline is adopted. 
\vspace{0.5em}
\vspace{0.5em}

\label{fig: sensitivity_tinyimgnet}
\newpage
\section{A Closer Look at the $\beta_1 = \beta_2$ Setting}
\label{sec: b1isb2}
\begin{figure}[t]
    \centering
\includegraphics[width=\linewidth]{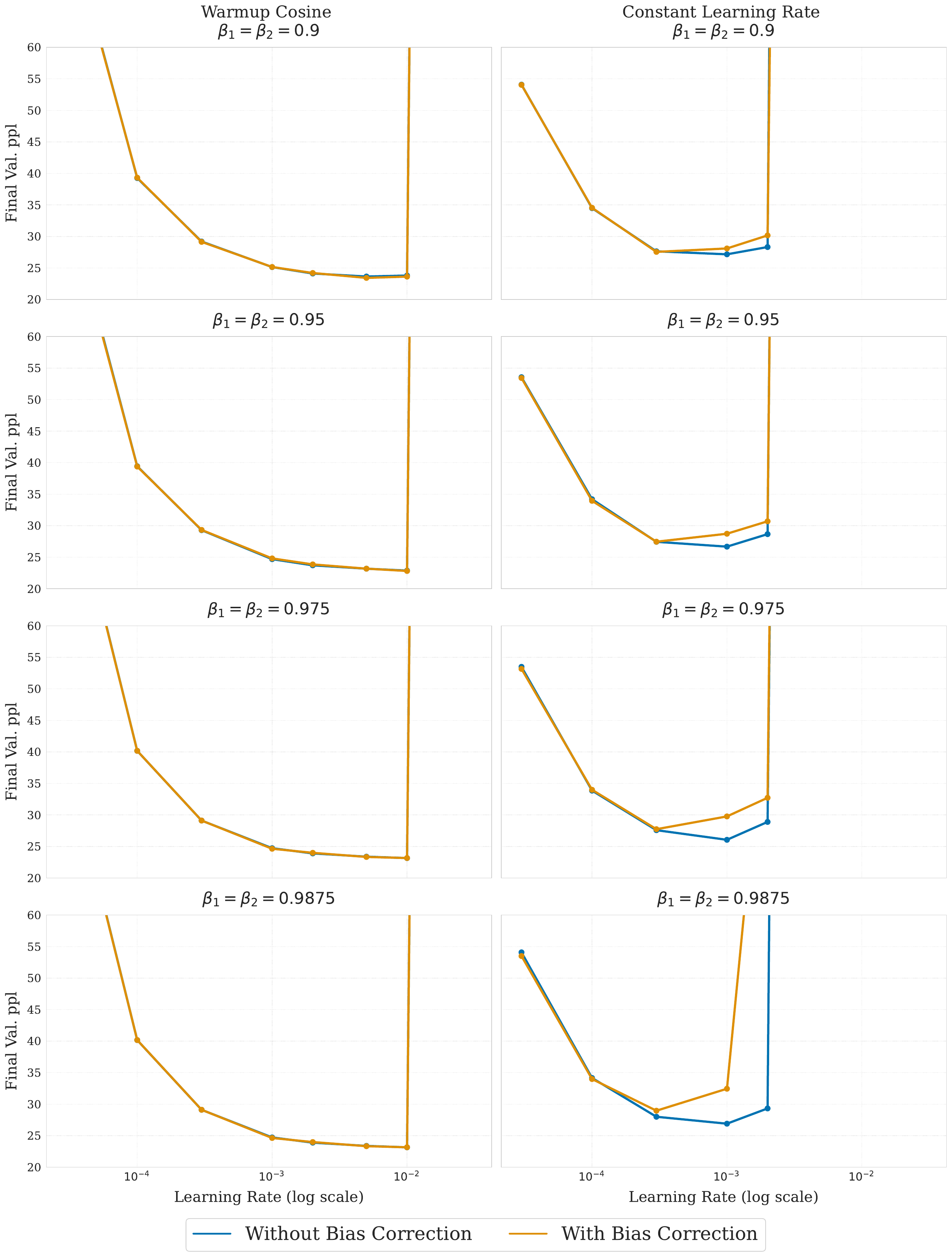}  
    \caption{
        Sensitivity curves comparing performance with and without bias correction (orange and blue respectively) for the case that $\beta_1 = \beta_2$.  The x-axis represents the learning rate (log scale) and the y-axis the final validation perplexity. \\
        With warm-up cosine scheduling, the inclusion of bias correction does not effect final validation perplexity across all learning rates.\\
        In contrast, with constant learning rate, bias correction actually denigrates performance, progressively more for larger values of $\beta_1 = \beta_2$ -- with bias correction always performing worse.
    }
    \label{fig: ppl_panel}
\end{figure}

\begin{figure}
    \centering
    \includegraphics[width=0.9\linewidth]{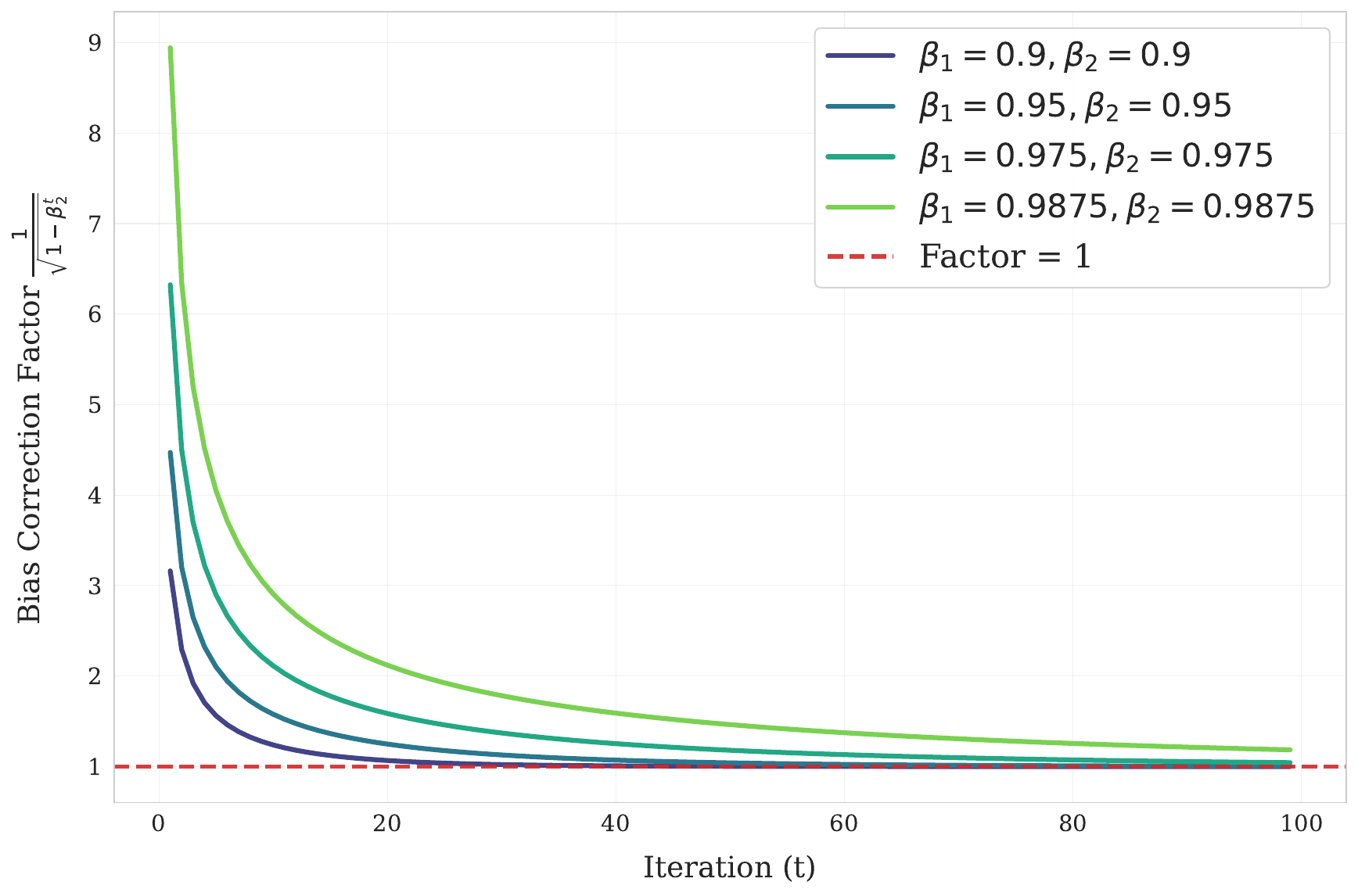}
    \caption{Plot of the \textit{bias correction} factor the first $100$ steps for $\beta = \beta_1 = \beta_2$\\
    As we can see, the behaviour for different $\beta$ values is similar, but larger values require more iterations to decay to $1$.} 
    \label{fig: b1_is_b2}
\end{figure}

We further investigate the final validation performance when considering different values of $\beta = \beta_1 = \beta_2$. We conduct a sweep across values $\beta \in \{ 0.9, 0.95, 0.975, 0.9875\}$ with the other training parameters the same as before. Due to computational constraints, each point represents a single random seed.\\
Consider Figure \ref{fig: ppl_panel}.When warm-up cosine scheduling is applied, final validation performance is indistinguishable for every learning rate and every choice of $\beta$.
On the other hand, when the learning rate is constant, the inclusion of bias correction always worsens optimal performance. Moreover increasing values of $\beta$ result in an increasing discrepancy between performance. 
\\
We posit that the performance discrepancy is explained by the behaviour of the bias correction factor as shown in Figure \ref{fig: b1_is_b2}. Indeed for higher values of $\beta$, the bias correction factor decays more slowly  towards the baseline. This spike in effective learning rate in the early steps likely has a destabilizing effect, resulting in a worse trained model.

\end{document}